\documentclass[letterpaper, 10 pt, conference]{ieeeconf}  

\IEEEoverridecommandlockouts                              

\usepackage{amsmath} 
\usepackage{amssymb}  
\usepackage{graphicx}
\usepackage{cite}

\usepackage{hyperref}

\usepackage{multirow}
\usepackage{threeparttable}
\usepackage{makecell}

\usepackage{booktabs}

\title{\LARGE \bf
ManiPose: A Comprehensive Benchmark for Pose-aware Object Manipulation in Robotics
}

\author{Qiaojun Yu$^{1*}$, Ce Hao$^{2*}$, Junbo Wang$^{1}$, Wenhai Liu$^{1}$, Liu Liu$^{3}$, Yao Mu$^{4}$, \\ Yang You$^{5}$, Hengxu Yan$^{1}$, and Cewu Lu$^{1\dagger}$
\thanks{$^{*}$ Equally contributed. $^\dagger$ Corresponding author.}
\thanks{$^{1}$Qiaojun Yu, Junbo Wang, Wenhai Liu, Hengxu Yan, Cewu Lu are with Department of Computer Science and Engineering, Shanghai Jiao Tong University, China, {\tt\small \{yqjllxs, sjtuwjb3589635689, sjtu-wenhai, hengxuyan, lucewu\}@sjtu.edu.cn }}%
\thanks{$^{2}$Ce Hao is with School of Computing, National University of Singapore, Singapore, {\tt\small cehao@u.nus.edu}}%
\thanks{$^{3}$Liu Liu is with Department of Computer Science and Information Engineering, Hefei University of Technology, China, 
{\tt\small liuliu@hfut.edu.cn}}%
\thanks{$^{4}$Yao Mu is with Department of Computer Science, The University of Hong Kong, China., {\tt\small muyao@connect.hku.hk}}%
\thanks{$^{5}$Yang You is with Department of Computer Science, Stanford University, USA, {\tt\small yangyou@stanford.edu}}%
}

\begin{document}

\maketitle
\thispagestyle{empty}
\pagestyle{empty}

\begin{abstract}
Robotic manipulation in everyday scenarios, especially in unstructured environments, requires skills in pose-aware object manipulation (POM), which adapts robots' grasping and handling according to an object's 6D pose. Recognizing an object's position and orientation is crucial for effective manipulation. 
For example, if a mug is lying on its side, it's more effective to grasp it by the rim rather than the handle.
Despite its importance, research in POM skills remains limited, because learning manipulation skills requires pose-varying simulation environments and datasets. This paper introduces \textit{ManiPose}, a pioneering benchmark designed to advance the study of pose-varying manipulation tasks. ManiPose encompasses:
1) Simulation environments for POM feature tasks ranging from 6D pose-specific pick-and-place of single objects to cluttered scenes, further including interactions with articulated objects.
2) A comprehensive dataset featuring geometrically consistent and manipulation-oriented 6D pose labels for 2936 real-world scanned rigid objects and 100 articulated objects across 59 categories.
3) A baseline for POM, leveraging the inferencing abilities of LLM (e.g., ChatGPT) to analyze the relationship between 6D pose and task-specific requirements, offers enhanced pose-aware grasp prediction and motion planning capabilities.
Our benchmark demonstrates notable advancements in pose estimation, pose-aware manipulation, and real-robot skill transfer, setting new standards for POM research. We will open-source the ManiPose benchmark with the final version paper, inviting the community to engage with our resources, available at our website \href{https://sites.google.com/view/manipose}{https://sites.google.com/view/manipose}.
\end{abstract}

\section{Introduction} \label{Sec: Intro}

\begin{figure}[t] 
    \centering
    \includegraphics[width=0.94\columnwidth]{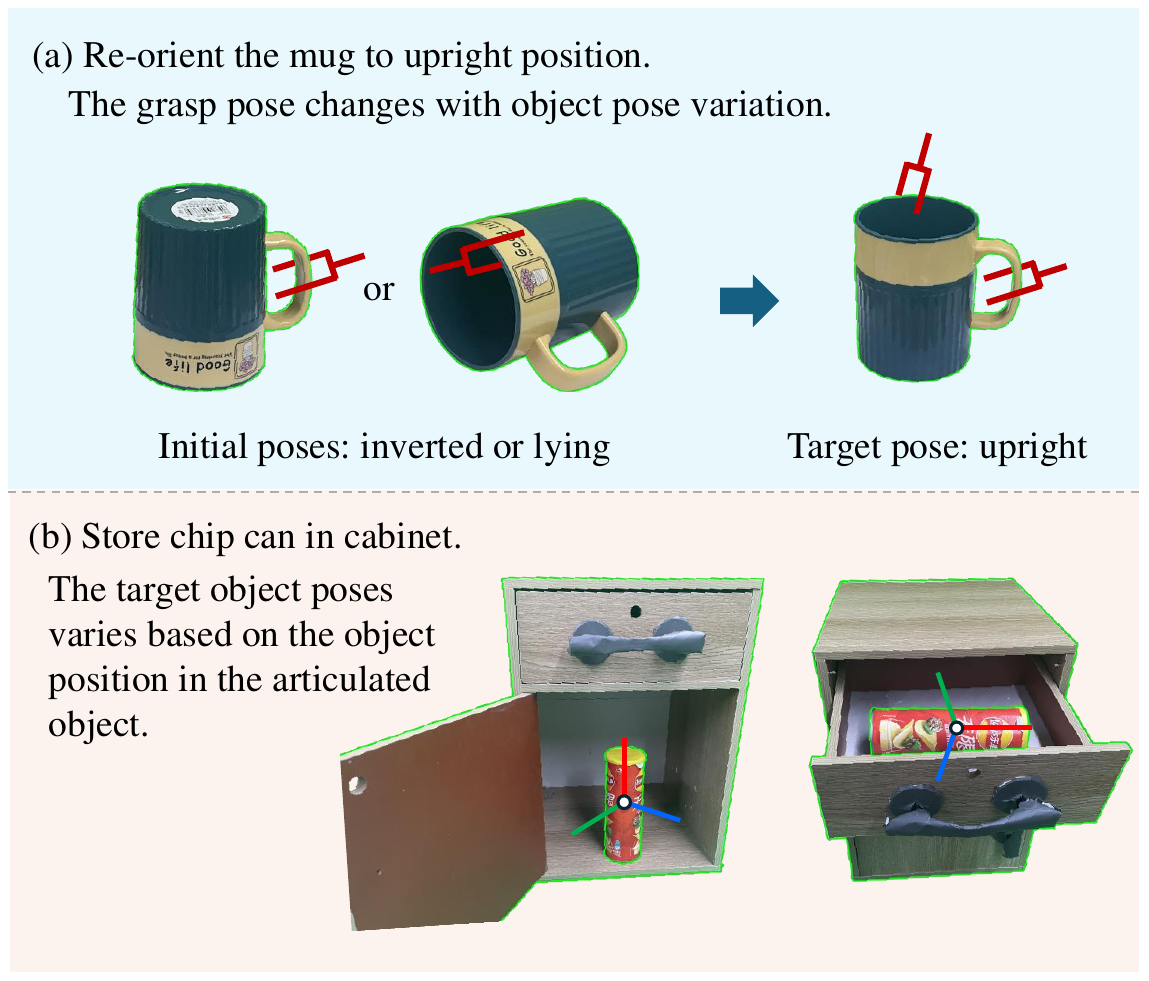}
    \caption{Illustration of the influence of object poses for manipulation.}
    \label{Fig: teaser}
\end{figure}

With the rapid advancement of multi-modal large models and their deep integration into the robotics field, robot manipulation skills for everyday scenarios have seen significant progress~\cite{brohan2023rt, driess2023palm, chi2023diffusion}. Different from traditional methods that directly learn a skill in an end-to-end manner, recent advances leverage multi-modal large models to interpret task descriptions and observe scenarios~\cite{kroemer2021review}, subsequently applying manipulation skills such as object grasping and trajectory planning to achieve their goals. For effective manipulation under this paradigm, it is crucial to understand the relationship between an object's 6D pose and the manipulation task, allowing the agent to dynamically adapt robots' grasping and handling according to an object's 6D pose. 
To demonstrate the impact of object pose on manipulation tasks, we provide two examples in Fig.~\ref{Fig: teaser}. In task (a), the goal is to re-orient a mug to an upright position from either an inverted or lying position. While the handle is the optimal grasp point for re-orienting an inverted mug, this approach is infeasible for a mug lying on its side due to its pose, making grasping the rim a more effective strategy. Task (b) involves storing a chip can in the cabinet drawer or door, where the target pose varies depending on the object's position relative to surrounding obstacles, highlighting the importance of pose information for adaptive manipulation strategies. We term this approach to manipulation, which actively incorporates object pose information, as pose-aware object manipulation (POM). 

POM is increasingly prevalent in household and assistant robots, reflecting the vast range of object poses encountered in daily life. Nonetheless, its adoption is limited in unstructured environments and everyday scenarios, largely due to the lack of comprehensive benchmarks and datasets specifically designed for pose-varying manipulation tasks. 
This paper aims to catalyze the development of learning algorithms for POM. To support this goal, we introduce ManiPose, a novel benchmark specifically designed for pose-aware manipulation tasks. ManiPose addresses the need for versatile simulation environments to test manipulation strategies and a large-scale object pose dataset for training foundational pose estimation modules.

ManiPose distinguishes itself through three principal features. 
First, we present benchmark simulation environments tailored for POM, featuring tasks that progress from pick-and-place of single objects with pose variations to operations in cluttered scenes, culminating in advanced interactions with articulated objects. The dynamic poses of objects introduce additional complexity, necessitating adaptable manipulation strategies informed by 6D pose. Second, our dataset, to our knowledge, is the largest of its kind, offering geometrically consistent, manipulation-oriented 6D pose labels for 2,936 real-world scanned rigid objects and 100 articulated objects across 59 categories. We adopt a uniform pose labeling convention, categorizing rigid objects into axis-symmetric, mirror-symmetric, and function-specific types, sourced from YCB~\cite{calli2015ycb}, OmniObject3D~\cite{wu2023omniobject3d}, and PACE~\cite{you2023pace}. Additionally, we provide carefully aligned labels for the joint and handle poses of 100 articulated objects in the categories of cabinets, ovens, and microwaves from PartNet-Mobility~\cite{xiang2020sapien}, ensuring consistent 6D pose definitions with function-specific labels and accurate scaling to match their real-world sizes.
Finally, we introduce a baseline method for POM, capable of generating viable grasp poses and trajectories based on the objects’ 6D pose, thereby offering a reference for future research in POM.

Moreover, we evaluate the ManiPose benchmark through three applications. Utilizing the CPPF algorithm~\cite{you2022cppf} for 6D pose estimation, we demonstrate enhanced accuracy and generalizability with our unified labeling approach. The baseline algorithm, when enhanced with ChatGPT4's inference and utilization of 6D pose information in POM environments, achieves effective manipulation. Furthermore, we successfully transitioned these algorithms from simulated environments to real-world robot applications, achieving comparable success.
 
In summary, the ManiPose benchmark is proposed to promote advancements in pose-aware manipulation. The subsequent sections of this paper are structured as follows: Section~\ref{Sec: related} reviews related work in POM and pose estimation. Sections~\ref{Sec: Task}, \ref{Sec: data}, and \ref{Sec: control} introduce the POM simulation environments, pose estimation dataset, and POM baseline, respectively, comprising the core components of ManiPose. Section~\ref{Sec: application} showcases applications of ManiPose in both simulated POM tasks and real-robot implementations. We conclude with an overview of the ManiPose benchmark and directions for future research.

\section{Related Works} \label{Sec: related}

\subsection{Robot Simulations and Manipulation Benchmarks}
Robot physics simulation and manipulation benchmarks are crucial for learning manipulation skills via imitation learning, reinforcement learning, and task and motion planning. Advanced physics simulation environments and Benchmarks such as CoppeliaSim (RLBench~\cite{james2020rlbench}), Mujoco~\cite{todorov2012mujoco} (robosuite~\cite{zhu2020robosuite}), PyBullet~\cite{coumans2021pybullet}, Isaac Gym~\cite{makoviychuk2021isaac} and Sapien~\cite{xiang2020sapien} (ManiSkill2~\cite{gu2023maniskill2}) provide high-fidelity dynamic robot simulation. These platforms simulate interactions with rigid-body, articulated, and soft-body objects, providing a rich array of observations including states of robots and objects, multi-view RGB-D images, and point clouds. 
Despite these advancements, a gap remains in these benchmarks' ability to simulate complex object poses and interactions accurately, limiting their effectiveness for developing pose-aware object manipulation (POM) strategies. ManiPose is designed to fill this gap by enhancing the Sapien simulation with explicit 6D object pose observation and introducing POM benchmark environments that feature varied object poses from isolated items to cluttered scenes and interactions with articulated objects. These enhancements aim to provide a robust development and validation tool for POM research, addressing the nuances of real-world object manipulation.

\subsection{Object 6D Pose Estimation and Datasets}

The field of object 6D pose estimation has evolved from focusing primarily on 2D image processing challenges to incorporating advanced techniques that leverage 3D data~\cite{guan2024survey}. Initiatives like PoseCNN~\cite{xiang2017posecnn} have explored predicting object poses by estimating their translation and rotation relative to the camera's viewpoint.  Approaches such as the normalized object coordinate space (NOCS)~\cite{wang2019normalized} have aimed at category-level 6D pose and size estimation from RGB-D images. Recent methodologies, including promptable object 6D pose estimation (POPE)~\cite{fan2023pope}, have further pushed the boundaries by enabling zero-shot open estimation using prompts. Concurrently, the development of 6D pose estimation datasets has enriched the domain, offering images and videos categorized at various levels of granularity. 

However, image-based 6D pose estimation methods often encounter challenges related to distribution shifts and noise from multi-view cameras, motivating point cloud as a more reliable data source for pose estimation. CPPF~\cite{you2022cppf, you2022cppf++} utilized point pair features (PPF) for category-level object 6D pose estimation from point clouds, but creating 3D object point clouds necessitates specific object meshes marked with poses. FoundationPose~\cite{wen2023foundationpose} and PACE~\cite{you2023pace} have made progress in collecting instance and category-level object meshes for point cloud-based 6D pose estimation. Despite these advancements, the lack of unified pose labeling across different categories has impeded the development of generalizable 6D pose estimation methods. To deal with the challenges, ManiPose offers an integrated dataset that incorporates real-world scanned rigid objects from diverse sources such as YCB~\cite{calli2015ycb}, OmniObject3D~\cite{wu2023omniobject3d}, PACE~\cite{you2023pace}, and articulated objects from PartNet Mobility~\cite{xiang2020sapien}, standardizing pose labeling across 3,036 objects in 59 categories to support higher-level pose estimation research.


\begin{figure*}[t]
    \centering
    \includegraphics[width=0.8\textwidth]{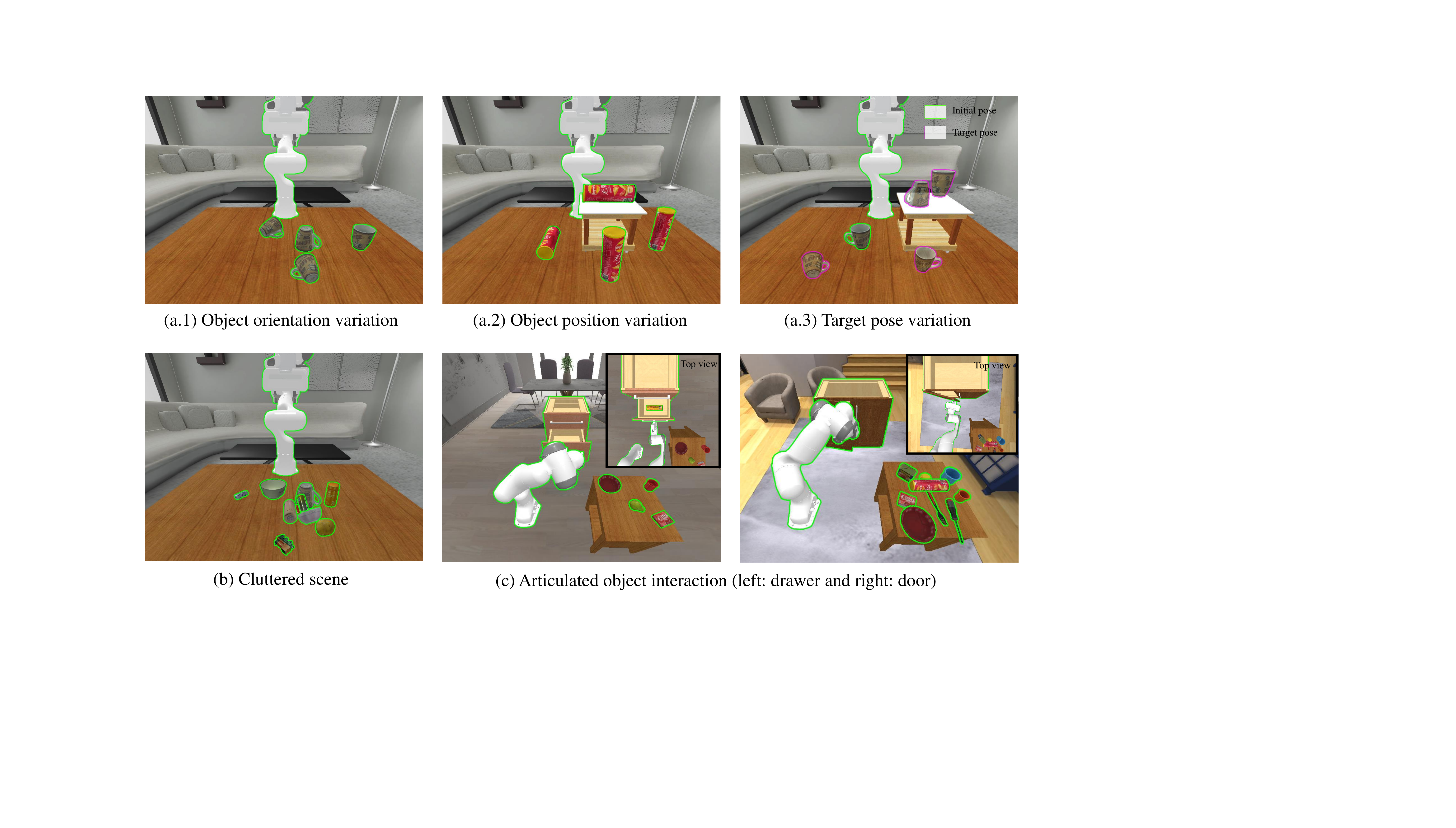}
    \caption{Pose-aware manipulation environments in ManiPose benchmark. \textbf{(a) Single object with pose variation}: Pick an object on the table and place it to a target pose. The initial and target poses (positions and orientations) are randomly selected. \textbf{(b) Multi objects in cluttered scene}: a pile of objects are place on the table. Pick one or more objects and place to target poses. \textbf{(c) Articulated object interaction}: Put in or take out objects to the cabinet drawer or door. The robot arm needs to open and close the articulated object by the handle and consider the relative poses of objects inside the cabinet.}
    \label{Fig: tasks}
\end{figure*}

\section{Pose-aware Manipulation Environments} \label{Sec: Task}

To evaluate the manipulation capabilities of controllers across varying object poses, the ManiPose benchmark has designed and created environments that are sensitive to the manipulation skills required for different object poses. Built on the physical simulator Sapien~\cite{xiang2020sapien}, these environments feature a variety of object combinations and pose variations to ensure that the learned manipulation skills are sufficiently generalizable. Within the Sapien environment, observations can be made of the object's state, as well as through 2D multi-view RGB-D images and point clouds. In the ManiPose environment, we have expanded the observation capabilities to include each object's pose information, allowing for a more comprehensive understanding of the scene. The definition of pose information for different object types and states varies according to the task, which will be detailed in Section~\ref{Sec: data}.
In terms of controlling the robot arm and gripper actions, ManiPose inherits all control modes from Sapien and introduces a long-horizon action primitive control command. This command enables movement based on the target's grasp pose and obstacles, with the specific methods to be discussed in Section~\ref{Sec: control}.

The pose-aware manipulation environments in ManiPose are illustrated in Fig.~\ref{Fig: tasks} and categorized into three main types: single object with pose variation, pick objects in cluttered scenes, and interact with articulated objects. 
In the latter part of this section, we will delve into every task and its variations, explaining the scenarios, functionalities, and the impact of different object poses on manipulation.

\subsection{Single object with pose variation} \label{Subsec: single}

In this environment, one object is placed on the table with randomized initial 6D poses (orientations and positions). The task is to grasp the object and move it to a target 6D pose with the robot arm and gripper. This task is different from the traditional pick-and-place environment that only randomizes the initial and target positions in a small range. Instead, we vary the initial 6D pose of the objects that can cause significant changes in the manipulation strategy (e.g. grasp pose and long-horizon trajectory). In addition, we control the target 6D pose for the object relative to obstacles, which also significantly influences the manipulation strategy. In the following, we specify three variations in the environment. 

\textbf{Object orientation variation:} 
The object is placed on the table and its initial orientation is randomly selected. For instance, in Fig.~\ref{Fig: tasks} (a.1), a mug on the table has four unique initial orientations. When grasping the mug, the object pose provides crucial information to determine the grasp point and approaching direction of the gripper on the mug. When the mug stands on the table, the handle can be easily grasped for manipulation. However, when the mug lies or leans on the table, the handle is not accessible by the gripper, so the grasp point should be changed to the rim or other regions. Therefore, the object orientation variation asks for the pose-specific awareness of the grasping actionability.

\textbf{Object position variation:}
The object is placed on the table with static obstacles and the object's position is randomly selected; therefore the grasping and manipulation behaviors are constrained by the object's relative position with the obstacle. For instance, in Fig.~\ref{Fig: tasks} (a.2), a chip can stands or lies near a shelf. Considering the relative position, the shelf limits the feasible grasping point and direction of the chip can. The object's position determines and modifies the manipulation decision.

\textbf{Target pose variation:}
The target pose (position and orientation) of the object is randomly selected in the 6D space. The task is successfully finished when the objects' 6D pose difference to the target is less than a small threshold. Although arbitrarily placing the object in obstacle-free space is easy, placing the object at a specific pose near other obstacles could be complex and long-horizon. For instance, in Fig.~~\ref{Fig: tasks} (a.3), the target pose of a mug is randomly scattered near a shelf, which may block and constrain the grasping and placing. Therefore, the manipulation should actively consider the target pose.

\subsection{Multi objects in cluttered scene}

In single-object manipulation environments, the objects do not have direct contact with other objects. However, in real-world scenarios, the objects being manipulated are often influenced by surrounding objects. Due to the spatial and angular relationships between multiple objects, the corresponding manipulation methods need to be adjusted. To study and test more generalized POM skills, we developed environments with multi-object cluttered scenes. As illustrated in Fig.~\ref{Fig: tasks} (b), a pile of objects is randomly placed on the table. Some objects have contact with each other, which leads to unpredictable object positions and orientations. The task is to move one or more specific objects to the target poses under the influence of other obstacles.

However, manipulating multiple objects is more complex than manipulating a single object. Contact between objects leads to stacking and occlusion, necessitating changes in the original manipulation strategies. For instance, the locations where objects can be grasped and the trajectories of their movement are restricted. Additionally, occlusion between objects can result in partial observations, introducing errors and uncertainties into pose estimation. These issues demand that manipulation capabilities possess sufficient understanding and generalization abilities. We design the environment with two challenges.

\textbf{Cluttered Objects variation:}
In various scenarios, multiple objects come into contact, imposing constraints on how objects are grasped and moved. As the types, positions, orientations, and relational dynamics of objects change, the manipulation and grasping of objects must also vary. This requires manipulation skills to have strong generalization abilities, i.e., the capacity to comprehend and plan accordingly in complex and variable scenarios.

\textbf{Object Occlusion and Stacking:}
In real-world scenes, occlusion and stacking among multiple objects can prevent accurate acquisition of object information, leading to errors in estimated poses. Solving this problem requires improvements in two aspects. The first is addressing the issue of partial observations through point clouds and images, which necessitates enhancing the accuracy of pose estimation via methods that integrate 2D and 3D information and dynamically update object information. The second aspect involves the manipulation and movement space constraints of occluded objects. The unobservable parts of an object may be in contact with other objects, necessitating long-horizon planning to comprehensively analyze the constraints present in manipulation.

\subsection{Articulated Object Interaction}
In conventional articulated object manipulation environments, the primary tasks involve directly opening or closing the articulated objects by the handle, coordinating the movement trajectory with the corresponding joint parameter. However, some types of articulated objects, such as drawers and doors, inherently serve the function of storing other objects. Therefore, an indispensable task within articulated object manipulation environments is the interaction process between portable objects and articulated ones. Moreover, during the manipulation process, the relative pose between objects and the articulated structures themselves significantly influences the decision-making for manipulation. 
As illustrated in Fig.~\ref{Fig: tasks} (c), in the environments, the robot needs to take objects out of a drawer or store objects in the cabinet. These tasks include the manipulation of articulated objects, (e.g. grasping handle and moving according to the joint axis), as well as interactions between portable objects and the storage containers. Since the cabinet has physical constraints, the manipulation must consider the possible object poses inside the container.

\section{Object Pose Estimation Dataset} \label{Sec: data}

\begin{figure*}[t]
    \centering
    \includegraphics[width=0.8\textwidth]{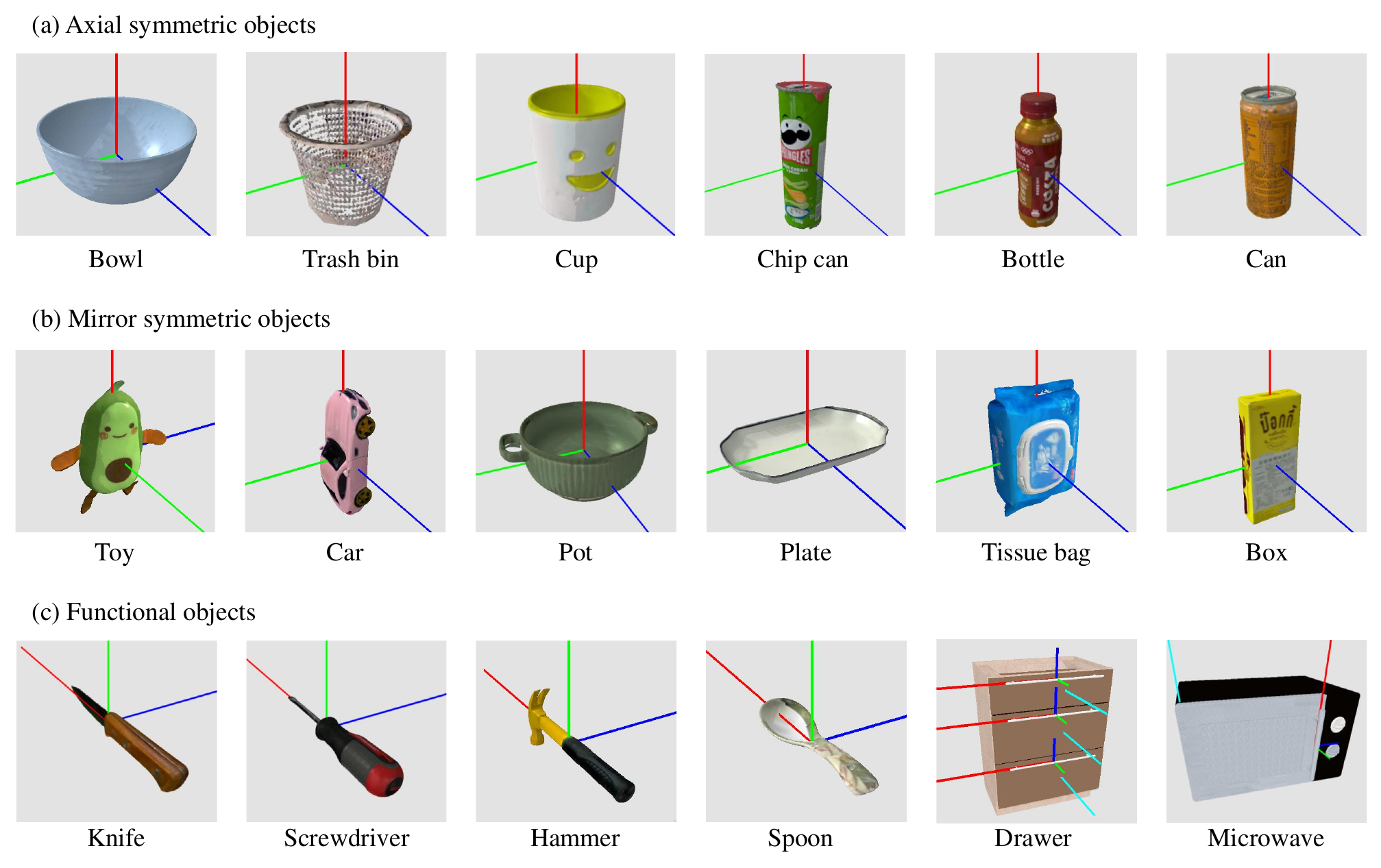}
    \caption{Object pose type-level alignment based on objects' geometry and functions. X, Y, and Z axes are represented in red, green, and blue. In the drawer and microwave, the joint axis is represented in cyan. \textbf{Axial symmetric objects}: X-axis is the axis of symmetry. \textbf{Mirror symmetric objects}: X-Y plane is the symmetric plane, and X-Z plane could be the second symmetric plane along a longer length. \textbf{Functional objects}: have functional and gripping areas. X-axis is the long axis direction; Y-axis is the grasp approaching direction.}
    \label{Fig: type_pose}
\end{figure*}

\begin{figure}[bth]
    \centering
    \includegraphics[width=0.8\columnwidth]{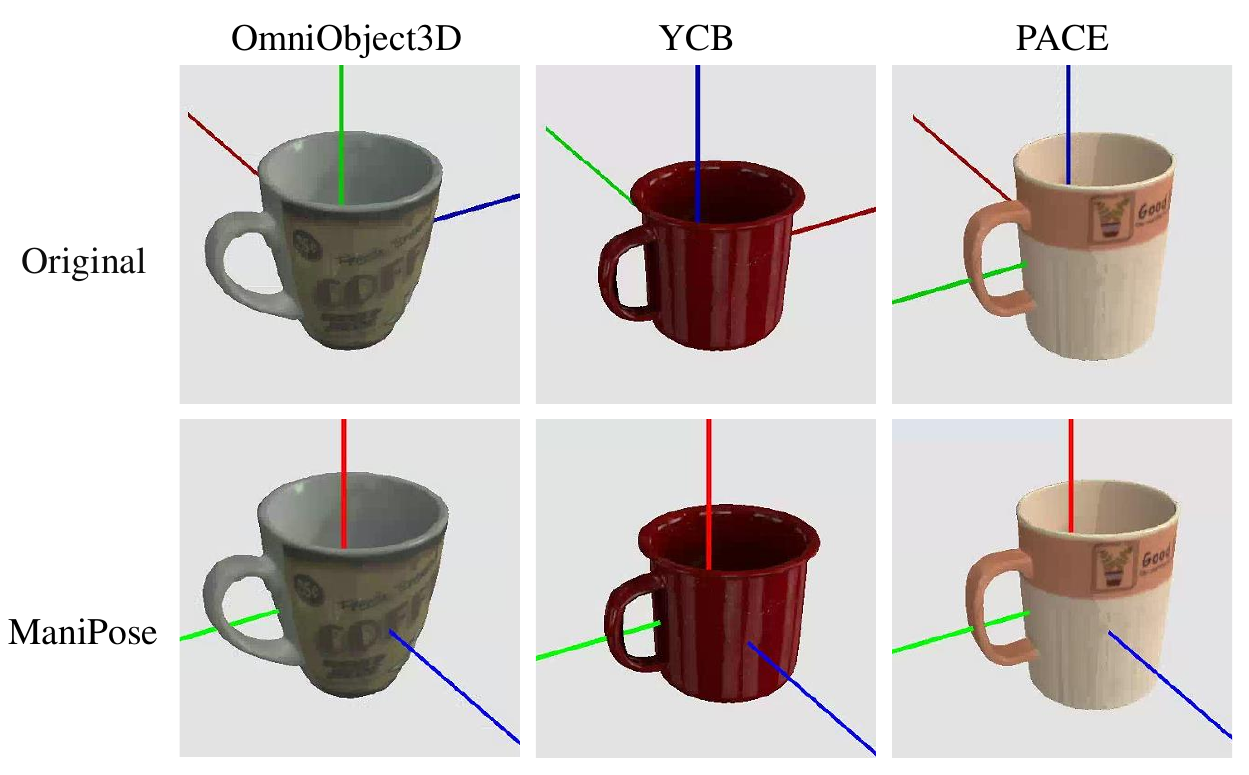}
    \caption{Object pose labeling. \textbf{Original}: mug pose labeling in OmniObject3D~\cite{wu2023omniobject3d}, YCB~\cite{calli2015ycb}, and PACE~\cite{you2023pace} datasets. \textbf{ManiPose}: unified mug pose labeling in ManiPose dataset. 
    X,Y, and Z axes are represented in red, green and blue. 
    }
    \label{Fig: pose diff}
\vspace{-3mm}
\end{figure}

In the realm of pose-aware object manipulation (POM), a fundamental task is observing and estimating the object poses. Traditional methods have been able to segment and track the 6D pose of objects from images and videos~\cite{wang2019normalized, xiang2017posecnn, sun2022onepose, fan2023pope}. However, these image-based recognition techniques often suffer from issues like camera noise and distribution shift, leading to inaccuracies, so they were confined to instance-level estimation. In contrast, advanced point cloud-based models~\cite{qi2017pointnet++} have been developed, enhancing recognition capabilities in 3D spaces and demonstrating robustness against environmental disturbances. Consequently, point cloud-based pose estimation methods~\cite{you2022cppf, you2022cppf++} have improved, advancing to general category-level estimation. Training these models necessitates object mesh datasets with accurately marked pose labels. Yet, existing datasets fall short in supporting the training of general cross-category pose estimation models suitable for large-scale POM tasks. A contributing factor is the misalignment of pose labeling across different categories and datasets. To mitigate this challenge, we have created a new object pose dataset in ManiPose, integrating object meshes from various sources and standardizing pose labeling based on object types. The ensuing sections delineate the dataset composition and the pose labeling methodology.

\subsection{General Object Pose Dataset} \label{Subsec: pose dataset}


Within ManiPose, we have established a comprehensive general object pose estimation dataset, integrating real-world scanned rigid objects from diverse repositories including YCB~\cite{calli2015ycb}, OmniObject3D~\cite{wu2023omniobject3d}, PACE~\cite{you2023pace}, and articulated objects form PartNet Mobility~\cite{xiang2020sapien}. The ManiPose dataset encompasses 3,036 objects across 59 categories. A notable addition to this dataset is the classification of functional objects, such as knives, toothbrushes, hammers, and screwdrivers. These objects have distinct pose requirements for effective grasping and manipulation, fulfilling specific functional purposes. Accurate pose estimation of these functional objects is crucial for facilitating their effective use in tool manipulation tasks.


Each object within the dataset is represented by a 3D mesh and associated texture, accompanied by a 6D pose label in Cartesian coordinates. The 6D pose encompasses the 3D translation from the base coordinate origin to the object’s pose coordinate system origin (position) and the rotation matrix representing the orientation from the base to the pose coordinate system. Visually, the axes of the pose coordinate system are color-coded in red, green, and blue, with the intersection point designated as the origin. Despite the original datasets having varied category-level object pose labels, leading to inconsistent pose estimation results, the ManiPose dataset addresses this issue. As illustrated in Fig.~\ref{Fig: pose diff}, we unify the pose labels for objects within the same category across different datasets, ensuring a standardized approach to pose estimation in the ManiPose dataset.

\subsection{Type-level Object Pose Alignment} \label{Subsec: pose align}



In previous pose estimation datasets, the pose estimation is implemented individually on every object instance. With the help of point cloud-based models~\cite{guan2024survey}, CPPF extended the pose estimation to the category level~\cite{you2022cppf}. In ManiPose, we further align the object pose labels with the same geometry and functional features as type levels. The type-level pose alignment can facilitate more general pose estimation training and provide meaningful pose information in the following pose-aware manipulation. In the following section, we classify the ManiPose dataset into three types in Fig.~\ref{Fig: type_pose}.

\textbf{Axial Symmetric Objects:}
In Fig.~\ref{Fig: type_pose} (a), the main parts of axial symmetric objects revolve around a rotation axis. For example, bowls, bottles, and cans are typical axial symmetric objects. The origin of the object's position is at the center of its bounding box, with the X-axis aligned with the rotation axis direction, typically vertically upward. The Y and Z axes are perpendicular to the X-axis, following the right-hand coordinate system constraint.

\textbf{Mirror Symmetric Objects:}
In Fig.~\ref{Fig: type_pose} (b), mirror symmetric objects have one or more symmetry planes. For instance, a car has one symmetry plane, a pot has two, and a box has three. The origin of the object's position is at the center of its bounding box. For objects with a single symmetry plane, the X-Y plane forms the object's symmetry plane, with the X-axis direction vertically upward, and the Y, Z-axes directions are determined by the right-hand rule. When an object has more than one symmetry plane, the X-Y and X-Z planes respectively form two symmetry planes. The X-axis direction is the same as with a single plane, but the Y-axis direction is along the longer length of the object, and the Z-axis direction is along the shorter length, forming a right-handed coordinate system. This definition utilizes the object's geometric shape to provide consistency in direction for objects with single or multiple symmetry planes.

\textbf{Functional Objects:}
In Fig.~\ref{Fig: type_pose} (c), functional objects, such as tools and handles on the articulated objects, play a crucial role in manipulation, where the grip and orientation significantly affect the tool's usage and the success of the task. Hence, functional objects' poses are separately labeled to facilitate task execution. 
The structure of tools is divided into the functional and grip areas. The functional area is the part of the tool that performs its function, and the grip area is where force and torque are applied for manipulation. For handles, the functional and grip areas are identical. 
The origin of the tool is defined as the joint point between the functional and grip areas. The origin of the handles is at the geometry center. The X-axis direction is along the long axis, and the Y-axis is perpendicular to the X-axis, indicating the direction from which the gripper approaches and grips the object. This labeling method explicitly defines the direction of the functional area, indicating the most suitable gripping direction for tool and handle usage during tasks.

\section{Pose-aware Manipulation Baseline} \label{Sec: control}

\begin{figure} [bht]
    \centering
    \includegraphics[width = 0.96 \columnwidth]{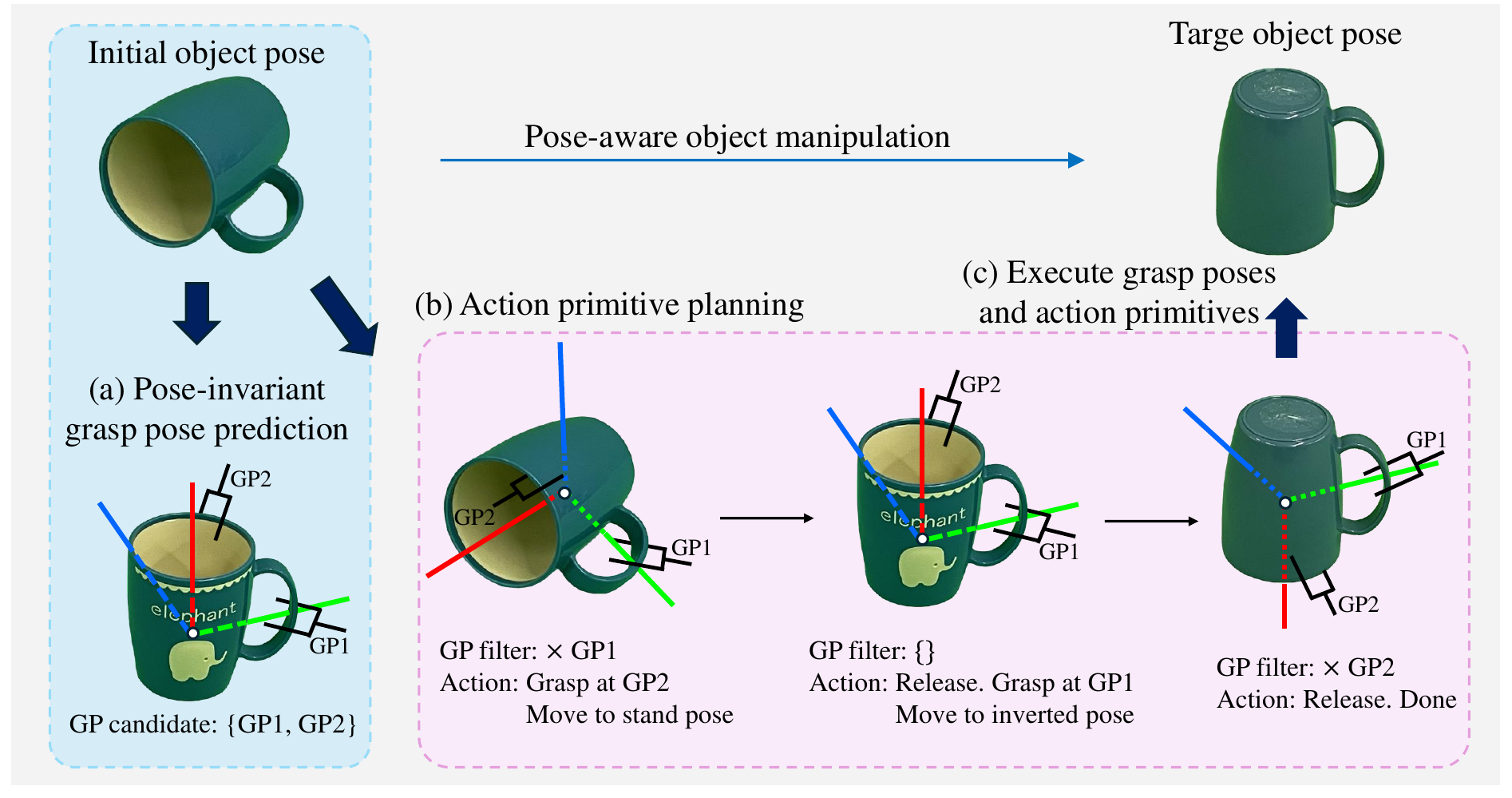}
    \caption{Pose-aware object manipulation baseline. \textbf{(a) Pose-invariant grasp pose prediction}: generate grasp pose (GP) candidates by converting objects to pose-invariant base coordinate. \textbf{(b) Action primitive planning}: plan trajectory with action primitives: \textit{Move to}, \textit{Grasp at}, and \textit{Release}. \textbf{(c) Execute} planned grasp poses and action primitives at each step.}
    \label{Fig: baseline}
\end{figure}

Traditional methods for learning manipulation skills, such as imitation learning, reinforcement learning, and random-shooting planning, focus primarily on two key tasks: grasp pose prediction and trajectory planning. Both tasks can be learned using data from simulators. However, as analyzed earlier, when there is variation in an object's initial and target poses, the manipulation strategy changes accordingly. Therefore, manipulation skills need to account for changes in object pose. In this section, we design a baseline method for POM, which generates successful trajectories in the ManiPose environments. This method provides a baseline structure for POM and generates demonstrations for imitation learning. The baseline method consists of two parts: pose-invariant grasp pose prediction and action primitive planning. We introduce the design in Fig.~\ref{Fig: baseline} and present experimental results in Section~\ref{Subsec: exp sim}.

\textbf{Pose-invariant grasp pose prediction.}
During the process of grasping an object, selecting an appropriate grasp pose is crucial. This can be achieved by predicting suitable grasp strategies through learning affordances~\cite{pohl2020affordance, varadarajan2011affordance}. However, when the object's pose changes and is influenced by obstacles, the grasp pose becomes constrained. Here, we propose a pose-invariant grasp pose (GP) prediction method in Fig.~\ref{Fig: baseline} (a). We first detect the object's pose using pose estimation methods and normalize the object to the base coordinate system to generate pose-invariant grasp pose candidates~\cite{fang2023anygrasp}. However, some grasp pose candidates are unreachable at unnormalized object poses, and we therefore apply a filter to rule out infeasible grasp poses under the physical constraints.

\textbf{Action primitive planning.}
When the objects' pose change cannot be finished in one grasping, the long-horizon manipulation could be challenging. For example, moving a lying mug to an inverted pose requires at least two grasping steps. Instead of planning the whole trajectory, we abstract the manipulation as high-level action primitives: \textit{Move to}, \textit{Grasp at}, and \textit{Release}. Given the task description and object poses, we use large language models like ChatGPT to reason and ground manipulation as action primitives~\cite{huang2023voxposer, brohan2023can}. In Fig.~\ref{Fig: baseline} (b), at each step, we filter grasp poses and plan action primitives to manipulate the mug to the target pose. Finally, we execute the planned grasp poses and action primitives to complete the task.



\section{Applications} \label{Sec: application}

\begin{figure*}[thb]
    \centering
    \includegraphics[width = 0.9 \textwidth]{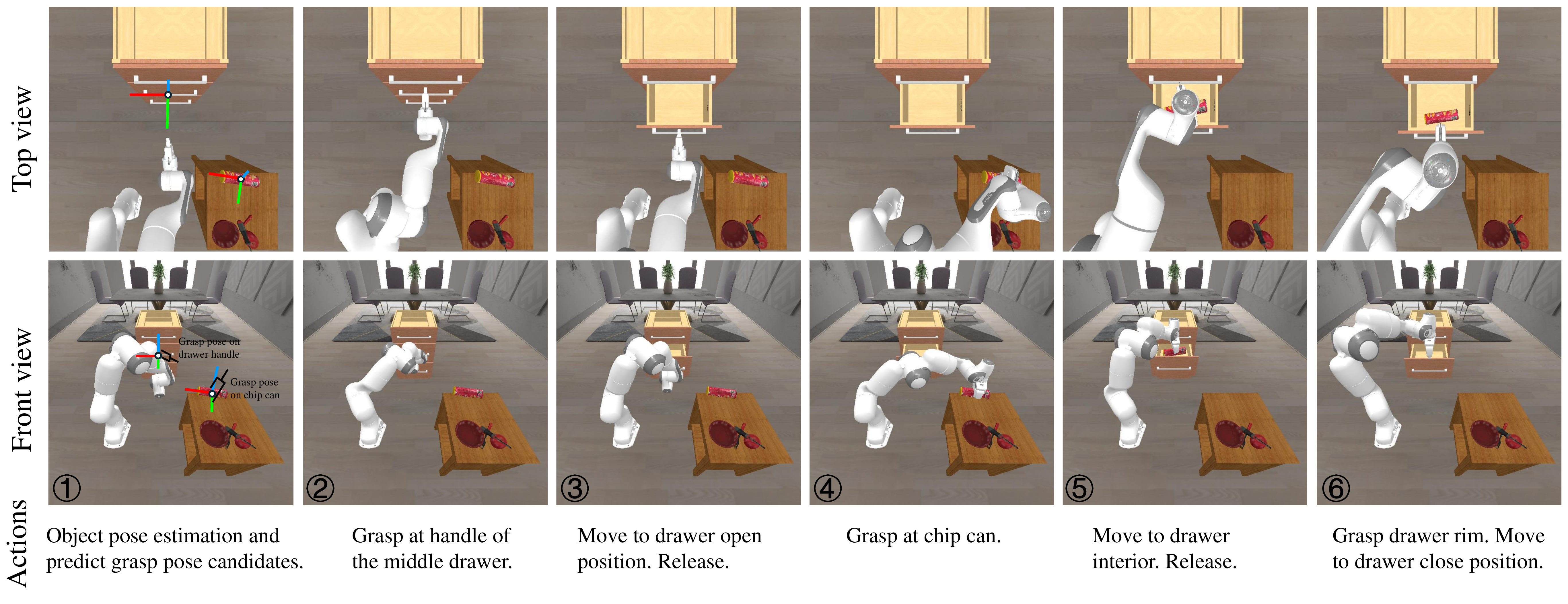}
    \caption{Demonstration of pose-aware object manipulation baseline. The task is to store the red chip can on the table in the middle drawer. In the initial step \textcircled{1}, the ManiPose baseline method estimates object poses, predicts pose-invariant grasp poses, and plans action primitives as in Section~\ref{Sec: control}. In the following steps \textcircled{2} to \textcircled{6}, the manipulation is sequentially executed.}
    \label{Fig: demo}
\vspace{-3mm}
\end{figure*}

In the previous section, we introduced the environments, object pose dataset, and POM baseline methods in ManiPose. In this section, we demonstrate the application methodologies and experimental results.
In POM tasks, the initial step is the pose estimation of the observed objects. We evaluated the point cloud-based pose estimation method, CPPF~\cite{you2022cppf, you2022cppf++}, as our baseline, conducting tests across various object types. 
Following the acquisition of object pose data, we applied the baseline POM method in section~\ref{Sec: control}, inferring manipulation strategies based on the pose information of objects. This approach yielded demonstrations in multi-object manipulation and interaction with articulated objects.
Finally, we apply the pose estimation and POM baseline from ManiPose to real-world robot experiments. The real-world experiments reveal and emphasize the importance of POM ability in everyday manipulation scenarios. 

\subsection{ManiPose Object Pose Estimation}

\renewcommand\arraystretch{1.3}
\begin{table}[ht!]
\caption{Pose Estimation Results}
\begin{center}
\begin{threeparttable}
\resizebox{\linewidth}{!}{
\begin{tabular}{c|l|cc|cc}
\toprule 

\multirow{2}{*}{Type$^1$} & \multirow{2}{*}{Category} & \multicolumn{2}{c|}{Origin error$^3$(cm) $\downarrow$} & \multicolumn{2}{c}{Orientation error$^3$($^\circ$) $\downarrow$}  \\

&  & Original$^2$ & ManiPose$^2$ & Original & ManiPose \\
\hline \hline

\multirow{2}{*}{Axial Symmetric}
  & Bowl & 1.517 & \textbf{1.231} & 37.853 & \textbf{5.097}  \\
  & Cup & 0.664 & \textbf{0.565} & 10.619 & \textbf{4.143}  \\

\midrule
\multirow{2}{*}{Mirror Symmetric}
  & Mug & 0.888 & \textbf{0.798} & 26.063 & \textbf{8.778}  \\
  & Box & 0.827 & \textbf{0.814} & 47.459 & \textbf{8.190}  \\

\midrule
\multirow{2}{*}{Functional Objects}
  & Hammer & \textbf{1.298} & 1.682 & 67.095 & \textbf{9.007}  \\
  & Razor & \textbf{1.139} & 1.203 & 59.885 & \textbf{12.019}  \\
  
\bottomrule
\end{tabular}}
\vspace{1.5mm}
\begin{tablenotes} 
\begin{minipage}{0.95 \columnwidth}
\item[1] Three object \textbf{types} defined in Section~\ref{Subsec: pose align} are axial symmetric, mirror symmetric and functional objects.
\item[$^2$] \textbf{Original}: object pose labeling from original datasets. \textbf{ManiPose}: unified object pose labeling in ManiPose (see Section~\ref{Subsec: pose dataset}).
\item[$^3$] Two evaluation metrics are: \textbf{Origin error}: pose origin position error with ground truth label and \textbf{Axis error}: pose orientation error with ground truth label.
\end{minipage}

\end{tablenotes}

\end{threeparttable}
\end{center}
\label{tab: pose}
\vspace{-10mm}
\end{table}

To validate the performance of type-level pose label alignment in ManiPose dataset (Section~\ref{Sec: data}), we employed point cloud-based pose estimation method CPPF~\cite{you2022cppf, you2022cppf++} for evaluation. CPPF conceptualizes pose estimation as a voting process, where it samples several point tuples from the point cloud. Each tuple then participates in voting for multiple target 6D poses, with the process culminating in the selection of the candidate possessing the highest vote count as the final pose estimate. In addition, a special vote selection scheme is applied to filter out some point tuples unsuitable for accurately predicting the voting target during inference. During training, an additional classification loss is applied to distinguish the good tuples, and the good tuples are determined by sorting the orientation loss.
In our experiments, we shuffled and divided the objects into $80\%$ training and $20\%$ testing sets, then rendered their corresponding point clouds. Subsequently, we trained CPPF on the training set for each category, using either aligned or original pose labels for supervision (Section~\ref{Sec: data}), and evaluated its performance on the testing set. We assessed the pose estimation by two metrics, objects' position (origin) error and orientation between the 
estimated pose and the labels~\cite{you2023pace}.


We show the pose estimation results in Table~\ref{tab: pose}. In all three types, the orientation errors were significantly reduced in the ManiPose dataset. This remarkable improvement is attributed to the type-level pose alignment to unify the objects across different datasets. In addition, ManiPose also slightly refined the origin errors of axial and mirror-symmetric objects, due to correct vote selection based on unified orientation. However, some performance degradation is observed in functional objects. We find a potential reason that these objects share similar geometry structures in the head and tail, and they are also relatively small and thin, which makes it challenging to accurately distinguish the origin and orientation.

\subsection{Experiments in ManiPose benchmark} \label{Subsec: exp sim}

We conducted comprehensive experiments with multi-object clutter and articulated object interaction in ManiPose environments in Section~\ref{Sec: Task} using the baseline method in Section~\ref{Sec: control}. As demonstrated in Fig.~\ref{Fig: demo}, the task is to store the red chip can on the task to the middle drawer. In the initial step \textcircled{1}, we observed the point cloud of every object and estimated the 6D pose of two relative objects, the chip can and the middle drawer, utilizing CPPF~\cite{you2022cppf} and GAMMA~\cite{yu2023gamma}. Then we employed  Anygrasp~\cite{fang2023anygrasp} to predict grasp poses on the chip can and drawer handle, based on the 6D pose, ChatGPT4  was then used to infer and select the appropriate grasp pose. Finally, we planned the trajectory by asking the ChatGPT4 to generate steps of action primitives given the task description~\cite{huang2023voxposer, brohan2023can}, and executed the actions in the following steps \textcircled{2} to \textcircled{6}. The demonstrations validate the effectiveness of ManiPose benchmark environments and POM baseline, with a success rate of $58\%$ across 100 trials.

\subsection{Real-robot Manipulation Experiments} \label{Subsec: exp real}

\begin{figure}[bth]
    \centering
    \includegraphics[width=0.9\columnwidth]{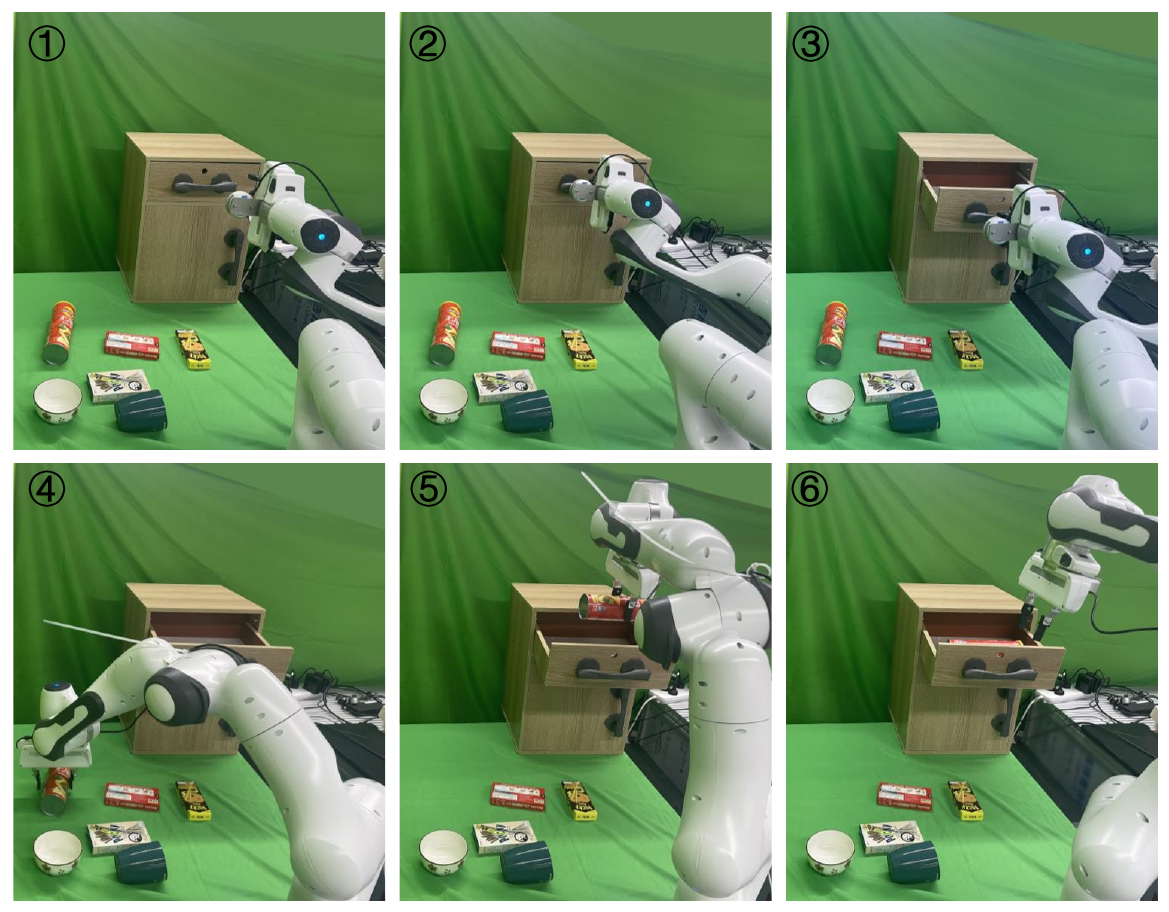}
    \caption{Transfer from ManiPose to real-robot experiments. The task is to store the red chip can in the drawer.}
    \label{Fig: real}
\vspace{-5mm}
\end{figure}

Following the experiments in ManiPose benchmark, we further transferred the POM manipulation skills to the real robot. In Fig.~\ref{Fig: real}, a Franka Emika Panda robot put the red chip can on the table into the drawer. With the exact object poses, the robot gripper accurately grasped the drawer handle and chip can and successfully manipulated the objects to the desired poses. The real-robot experiment showed a strong generalization ability of pose-aware object manipulation (POM) and transferability from simulation to the real world. 

\section{Conclusion}
In conclusion, this paper presents ManiPose, an innovative benchmark designed to enhance research in pose-aware object manipulation (POM), enabling more effective use of pose variations in robotic manipulation tasks. ManiPose encompasses simulation environments for POM, a comprehensive dataset with geometrically consistent
and manipulation-oriented 6D pose labels, and a baseline method leveraging LLM inferencing capabilities for improved grasp prediction and motion planning. Through extensive experiments, we demonstrated that ManiPose significantly advances pose estimation, POM, and real-robot skill transfer, thereby setting a new benchmark for robotic manipulation research. Future work will focus on refining the baseline methods and exploring additional applications to enhance the utility and applicability of ManiPose in the field of POM.

\bibliographystyle{IEEEtran}
\bibliography{reference}

\end{document}